\documentclass{article}

\date{July 25, 2025}

\usepackage{arxiv}

\usepackage[utf8]{inputenc}   
\usepackage[T1]{fontenc}      

\usepackage{amsmath}
\usepackage{amsfonts}

\usepackage{booktabs}         
\usepackage{tabularx}         
\usepackage{float}            

\usepackage{graphicx}         
\graphicspath{{media/}}       

\usepackage{microtype}        
\usepackage{nicefrac}         

\usepackage{listings}         
\usepackage[most]{tcolorbox}  
\usepackage{newclude}         

\usepackage{fancyhdr}         

\usepackage{tikz}
\usetikzlibrary{positioning, fit, calc}  

\usepackage{xurl}             
\usepackage{hyperref}         
\hypersetup{breaklinks=true}

\usepackage{enumitem}
\setlist[itemize]{topsep=6pt, itemsep=6pt} 

\renewcommand{\undertitle}{A Preprint}

\title{Efficient and Scalable Agentic AI with Heterogeneous Systems
}

\author{
  Zain Asgar \\
  Stanford University, Gimlet Labs Inc. \\
  Stanford, CA\\
  \texttt{zasgar@stanford.edu} \\
   \And
  Michelle Nguyen \\
  Gimlet Labs, Inc \\
  San Francisco, CA\\
  \texttt{michelle@gimletlabs.ai} \\
  \AND
  Sachin Katti \\
  Stanford University, Intel \\
  Stanford, CA \\
  \texttt{skatti@stanford.edu} \\
}

\begin{document}
\maketitle

\begin{abstract}
AI agents are emerging as a dominant workload in a wide range of applications, promising to be the vehicle that delivers the promised benefits of AI to enterprises and consumers. Unlike conventional software or static inference, agentic workloads are dynamic and structurally complex. Often these agents are directed graphs of compute and IO operations that span multi-modal data input and conversion (e.g. speech to text), data processing and context gathering (e.g privacy filtering, vector DB lookups), multiple LLM inferences, tool calls, etc. To scale AI agent usage, we need efficient and scalable deployment and agent-serving infrastructure. Today, however, the vast majority of these workloads are deployed on homogenous, high-end, single-vendor infrastructure, which can often be quite expensive and limits broad rollout.

To tackle this challenge, in this paper, we present a system design for dynamic orchestration of AI agent workloads on heterogeneous compute infrastructure spanning CPUs and accelerators, both from different vendors and across different performance tiers within a single vendor. The system delivers several building blocks: a framework for planning and optimizing agentic AI execution graphs using cost models that account for compute, memory, and bandwidth constraints of different HW; a MLIR based representation and compilation system that can decompose AI agent execution graphs into granular operators and generate code for different HW options; and a dynamic orchestration system that can place the granular components across a heterogeneous compute infrastructure and stitch them together while meeting an end-to-end SLA. Our design thus performs a systems level TCO optimization and our preliminary results show that leveraging a heterogeneous infrastructure can deliver significant TCO benefits. A preliminary surprising finding is that for some workloads a heterogeneous combination of older generation GPUs with newer accelerators (H100 and Gaudi 3 respectively) can deliver similar TCO as the latest generation homogenous GPU infrastructure design (such as clusters of B200s), potentially allowing us to leverage deployed GPU infrastructure for longer periods than previously assumed. 
\end{abstract}

\keywords{Agentic AI \and MLIR \and Heterogeneous Hardware \and Scheduling \and Compilers \and Systems for ML \and LLM Inference \and GPU \and CUDA}

\pagebreak
\section{Introduction}

Agentic AI is experiencing rapid growth, with market research indicating significant adoption across various industries. Recent surveys suggest that over 75\% of enterprises are actively deploying or evaluating agentic AI solutions due to their ability to augment or automate complex workflows \cite{gravitee2025,survey_cloudera2025,pwc2025}. This rapid adoption is driven by agentic AI's capability to integrate large language models (LLMs), multimodal models (e.g., speech, text, images), intricate data processing techniques, database queries, and external API integrations. Unlike traditional AI applications, which typically involve straightforward model serving scenarios, agentic AI dynamically orchestrates multiple models and heterogeneous tasks, creating complex execution patterns and interdependencies. Efficiently scaling infrastructure to support these multifaceted workloads is critical for unlocking agentic AI's full transformative potential, delivering clear operational benefits to both enterprises and consumers.

Prior work has extensively explored frameworks and techniques for building AI agents, aiming to enhance their accuracy, safety, and security. Notable examples include ReAct \cite{yao2023reactsynergizingreasoningacting}, AutoGPT \cite{yang2023autogptonlinedecisionmaking}, and CAMEL \cite{li2023camelcommunicativeagentsmind}. Researchers have also addressed significant risks related to security and safety, such as unpredictable user inputs and interactions with untrusted external systems \cite{deng2024aiagentsthreatsurvey}, and have proposed effective designs for multi-agent systems requiring standardized protocols \cite{dang2025multiagentcollaborationevolvingorchestration, yang2025surveyaiagentprotocols}. Despite these advances, research specifically focused on developing efficient and scalable infrastructure tailored for agentic AI workloads remains nascent.

While there has been limited research dedicated specifically to building scalable infrastructure for agentic AI, complementary efforts have extensively optimized individual model serving. Notable developments include kernel-level innovations like Flash Attention, significantly improving inference performance and memory efficiency for transformer models. Additionally, advanced optimization techniques such as prefill and decode disaggregation have emerged, allowing more efficient utilization of compute resources. Other significant research has explored operational efficiency through model optimization techniques like sparsity and quantization \cite{zheng2024learnefficientbuildstructured, li2024evaluatingquantizedlargelanguage}, optimized execution engines such as vLLM \cite{kwon2023efficientmemorymanagementlarge}, and dynamic batching methods \cite{Narayanan2021}. However, these approaches predominantly target static inference scenarios involving a single model and thus do not adequately address the dynamic nature of agentic AI workloads. Agentic AI execution consists of multiple, varying models and data processing tasks that interact dynamically, necessitating fundamentally new systems-level approaches tailored specifically to their inherent complexity and variability. Recent studies have further highlighted that assessing agents solely on accuracy can lead to complex and expensive workloads, underscoring the importance of cost-aware benchmarks \cite{kapoor2024aiagentsmatter}.

A central insight of this paper is that efficiently executing agentic AI workloads requires moving beyond traditional homogeneous GPU deployments to heterogeneous systems. These heterogeneous systems are composed of accelerators across different vendors and performance tiers within a single vendor. Our analysis shows that agentic AI workloads can be decomposed into granular components, each exhibiting sensitivity to distinct hardware resource specifications such as TFLOPS, memory bandwidth and capacity, network bandwidth, disk capacity, and general-purpose compute. By aligning these granular computational tasks—such as LLM prefill, LLM decoding, data processing, and API interactions—with specifically optimized hardware capabilities, we can significantly reduce the Total Cost of Ownership (TCO) and enhance the efficiency of AI inference deployments. 

However, exploiting this insight requires framing the solution space for building efficient infrastructure tailored specifically for agentic AI workloads. Effective AI infrastructure must be designed with the capability to maintain a holistic and granular view of the entire agentic pipeline, comprehensively understanding the interactions and dependencies between all its components. It must also have the flexibility to deconstruct pipelines into even more granular tasks. Furthermore, a dynamic, cost-aware orchestration layer is essential to intelligently place individual components onto suitable hardware resources, seamlessly integrate them, and optimize overall system cost. This orchestration must simultaneously ensure compliance with application-level Service Level Agreements (SLAs), such as user-perceived latency and throughput. Achieving these capabilities necessitates reimagining every layer of the AI infrastructure stack—from model runtimes and dynamic compilers to orchestration frameworks and observability solutions.

In this work, we propose a comprehensive systems-level approach designed explicitly for efficient execution of agentic AI workloads. Our contributions include:
\begin{itemize}
\item{\textbf{MLIR based Dynamic Dataflow Representation \& Compilation:} Modeling and compiling agentic workloads as dynamic, granular execution graphs leveraging an MLIR based toolchain, explicitly capturing the complexity and variability inherent to these workflows.}
\item{\textbf{Cost-Aware Optimization Framework:} Formulating an optimization strategy that schedules and executes agentic AI workloads efficiently under practical latency, throughput, and resource constraints.}
\item{\textbf{Heterogeneous Hardware Integration:} Incorporating heterogeneous hardware across different price points and vendors to leverage task-specific computational strengths, improving overall system efficiency and reducing cost.}
\item{\textbf{Dynamic AI Agent Orchestration:} Developing an agentic AI orchestrator that integrates these software- and hardware-level optimizations, enabling scalable, efficient, and cost-effective deployment of agentic AI systems at enterprise scale.}
\end{itemize}

\section{Defining AI Agent Workloads}

AI agents can broadly be defined as computational entities that perceive their environment, process information, and act toward a specified goal. Unlike traditional software governed by static control flow, agents operate based on data, models, and dynamic policies. Their design spans a wide spectrum, ranging from simple, rule-based systems to complex, autonomous frameworks capable of multi-step planning, memory management, and external tool integration.

\subsection{Autonomy and Complexity}
AI agents can differ significantly in their level of autonomy and internal complexity. At one end of the spectrum are reflex based agents that follow simple, condition-based logic such as spam filters or thermostats that map input to output using predefined rules \cite{russell1995artificial}. More advanced agents exhibit autonomy by reasoning over multiple steps and selecting actions through a closed loop of perception, decision-making, and tool use. For example, AI agents can be leveraged to answer open-ended questions with real data by dynamically querying external sources, such as Wikipedia or search APIs \cite{yao2023reactsynergizingreasoningacting}. 

These workflows demand richer architectural structure to effectively coordinate control flow and memory over time. As agent behaviors become more complex, systems have evolved from single agent pipelines to multi-agent architectures with distinct hierarchies. Figure \ref{fig:agent-architectures} presents a taxonomy of common agentic structures, illustrating how control can flow through single agents, coordinated peers, hierarchical layers, or fully custom graphs. For example, in a peer-to-peer agent network, multiple agents operate concurrently on different sub-tasks and collaborate by exchanging information to achieve a shared goal. In contrast, a hierarchical architecture introduces structured layers of control, where higher level agents handle planning and decision making, while lower level agents focus on specialized execution tasks.

\subsection{Interactivity}
Another dimension of variation is interactivity, which refers to how agents engage with their environment or users throughout task execution. Interactive agents operate in a feedback loop, adjusting behavior in response to real-time inputs. These agents often expose APIs, hold state across turns, and issue clarifying queries to disambiguate user intent. This design is well-suited for customer service bots, coding copilots, or multi-step reasoning tasks. In contrast, non-interactive agents execute in a single pass, producing outputs without runtime adaptation. Such agents are commonly used for document summarization, batch inference, or fixed automation pipelines. 

\subsection{Functional Capabilities}
Agents are equipped with a diverse set of functional capabilities, encompassing the concrete operations an agent can perform. Some operate in closed environments, relying solely on static model parameters to generate responses. Others can dynamically invoke tools, such as search engines, APIs, databases, or code execution environments, to augment their reasoning or access real time information. This tool-using behavior enables agents to go beyond static knowledge, making them more robust and useful in open ended or evolving environments.

\begin{figure}[ht]
    \centering
    \includegraphics[width=0.7\textwidth]{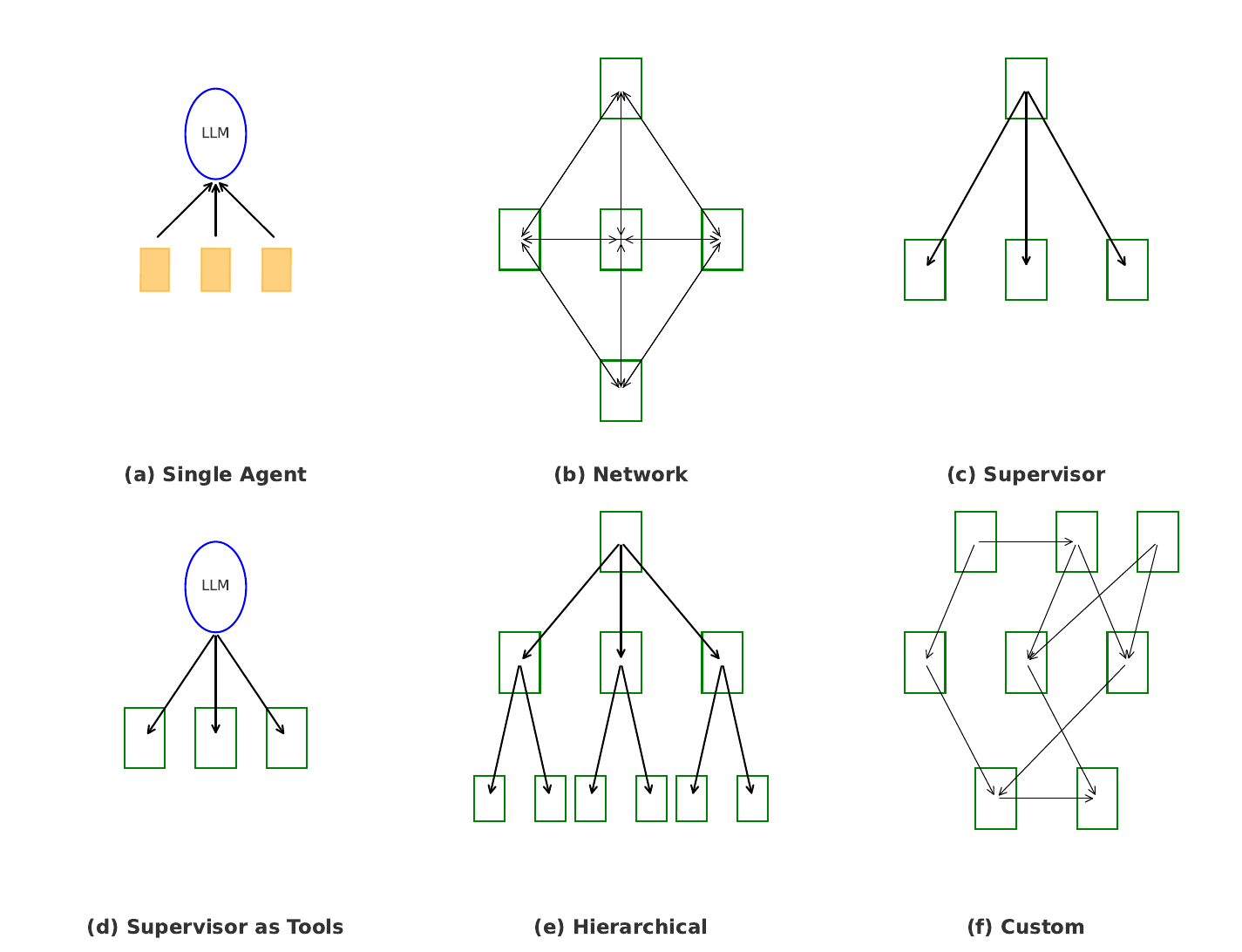}
    \caption{
    Comparison of agentic architectural patterns, inspired by LangGraph's taxonomy~\cite{langgraph2024multiagent}.
    (a) A single LLM agent invoking external tools directly. 
    (b) A peer-to-peer network of agents coordinating actions. 
    (c) A supervisor agent dispatching work to multiple subordinate agents. 
    (d) A single agent that uses another agent (e.g., a supervisor) as a tool. 
    (e) A hierarchical architecture with clear delegation across layers, a generalized version of the supervisor pattern. 
    (f) A custom, arbitrarily structured agent graph enabling flexible planning.
    }
    \label{fig:agent-architectures}
\end{figure}

\subsection{Agents as a compute graph}
To understand and optimize the behavior of AI agents, we need a representation that captures both their modular composition and dynamic control flow. Such a representation enables systematic analysis of execution dependencies, identification of bottlenecks, and opportunities for optimization and scheduling.

A natural way to express agent workloads is as a directed, potentially cyclic, graph of tasks. This graph represents the dataflow and execution dependencies between components. Each node in this graph corresponds to a discrete operation or module. These nodes are hierarchical, where the node may itself be an agent composed of further subgraphs. This allows us to represent agent workloads following the taxonomy of common agentic architectures in Figure \ref{fig:agent-architectures}.

We outline the tasks of the dataflow graph in detail in Table \ref{tab:agent-tasks}, where we provide example inputs, outputs, and implementations for each node type. At a high level, these nodes include:

\begin{itemize}
    \item \textbf{Agent:} A nested or composite controller with its own task graph.
    \item \textbf{Tool Call:} An external API or function invoked as part of execution.
    \item \textbf{Model Execution:} A transformer-based inference step.
    \item \textbf{Memory:} Access to external context or database.
    \item \textbf{General Purpose Compute:} Lightweight CPU-side processing for logic, parsing, or transformation.
\end{itemize}

While nodes in the agent graph represent discrete operations, the edges define the dataflow and control dependencies between those operations. An edge typically indicates that one node’s output is required by another, such as text, embeddings, tool outputs, or control signals. Edges can represent either synchronous or asynchronous execution. In conditional or cyclic graphs, they may also encode feedback loops or branching behavior.

\begin{figure}
\centering
\includegraphics[width=0.85\textwidth]{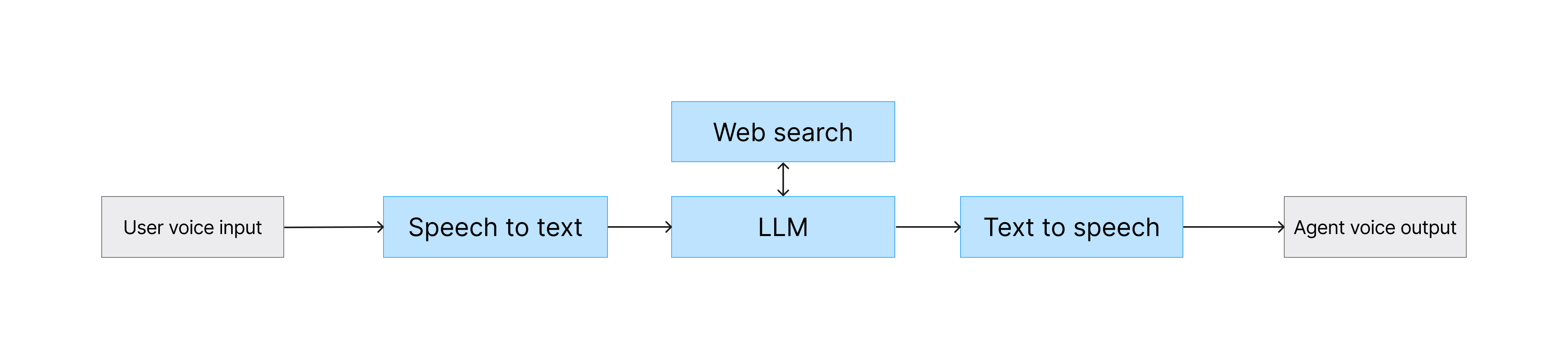}
\caption{Directed graph for a conversational voice agent}
\label{img:agent-dag}
\end{figure}

\subsubsection{Dataflow Graph Example}
To illustrate how a simple agent can be modeled as a dataflow graph, consider a conversational voice agent designed to answer user questions using web search. The graph begins with a user’s spoken query as the input node. This signal is transcribed using a Speech-to-Text model and passed to a language model for processing. If the LLM determines that additional context is needed, it triggers a branch that issues web search queries to retrieve relevant information. This process may repeat until the model has enough context to generate a complete response. The final output is converted to speech using a Text-to-Speech model and returned to the user. The complete computation graph, including conditional control flow, is shown in Figure \ref{img:agent-dag}.

\subsubsection{Systems-Level Optimizations of Dataflow Graphs}
Representing agents as dataflow graphs provides a natural abstraction for modeling the flow of information and computation across discrete operations. This structure not only makes agent behavior more interpretable and modular but also illustrates the space for system-level optimization such as parallelism. Since many agent workloads involve multiple independent or loosely coupled operations, graph structure can reveal natural concurrency across tasks or stages of execution. Leveraging this parallelism is critical for improving latency, throughput, and resource utilization. The primary forms of parallelism exposed by agent graphs include:
\begin{itemize}
\item \textbf{Pipeline parallelism}:  Decomposing a workload into sequential stages, where each stage processes a different input concurrently. This enables overlapping execution and improved throughput.
\item \textbf{Task parallelism}: Executing independent operations or sub-tasks in parallel, often across separate threads, processes, or nodes.
\end{itemize}

A concrete instance of pipeline parallelism is disaggregated inference, where LLM execution is partitioned into prefill and decode stages and scheduled across distinct hardware resources. This staged architecture enables overlapped execution, allowing the system to initiate prefill for a new input while performing decode on a prior output.

In contrast, an example of task parallelism is expert parallelism. In expert parallelism, different model components are specialized for specific sub-tasks and invoked concurrently. Each expert operates independently on its assigned portion of the workload, enabling the system to process multiple computations simultaneously.

\begin{table}[h]
\centering
\caption{Common Agent Task Types}
\label{tab:agent-tasks}
\begin{tabularx}{\textwidth}{l X X X}
\toprule
\textbf{Task Type} & \textbf{Inputs} & \textbf{Outputs} & \textbf{Example} \\ 
\midrule
Agent & Messages, context & Sub-tasks or output message & Recursive agent graph \\[16pt]
Model Execution & Token sequence & Token logits, hidden state & Llama \cite{touvron2023llamaopenefficientfoundation}, GPT \cite{radford2018gpt}, BERT \cite{devlin2019bertpretrainingdeepbidirectional} \\[16pt]
Model KV Cache & Write from prefill; Read from decode & KV tensors & Device-local memory, remote memory, or disaggregated store \\[16pt]
Tool Call & Query or structured input & Tool response or data & API request (e.g., calculator, search) \\[16pt]
Memory Lookup & Key, retrieval prompt & Retrieved documents or values & Vector DB (e.g., FAISS, PGVector) \\[16pt]
General Purpose Compute & Data blob, parameters & Transformed data & JSON parsing, routing logic \\[16pt]
Control Flow / Planner & Graph state, input tokens & Execution plan or subgraph & Agent planner module \\[16pt]
Observation Store & Event, result & Updated memory state & Logging, episodic memory \\
\bottomrule
\end{tabularx}
\end{table}

\subsection{Agent Task Workload Characteristics}

Each node in an AI agent's dataflow graph exhibits distinct workload requirements across a set of key hardware dimensions. For instance, a node performing LLM inference may be GPU-bound, with high demands on memory capacity and floating-point throughput. Meanwhile, a tool invocation node is likely to be I/O-bound and dominated by network latency.

Concretely, we define these key hardware dimensions as:
\begin{itemize}
\item \textbf{High Performance Compute}: Captures the ability of specialized hardware, such as GPUs or AI accelerators, to run compute-intensive operations with high FLOP requirements.
\item \textbf{Memory Bandwidth}: Measures the rate at which data can be read from or written to memory.
\item \textbf{Network Bandwidth}: Captures the capacity to move data across nodes or services. High network bandwidth supports low latency communication between distributed components.
\item \textbf{Memory Capacity}: Refers to the total available memory on a device or system.
\item \textbf{Disk Capacity}: Measures the amount of persistent storage available.
\item \textbf{General Purpose Compute}: The ability to process scalar CPU-based operations, such as logic, parsing, orchestration, or preprocessing.
\end{itemize}

\begin{figure}[H]
    \centering
    \includegraphics[width=0.75\textwidth]{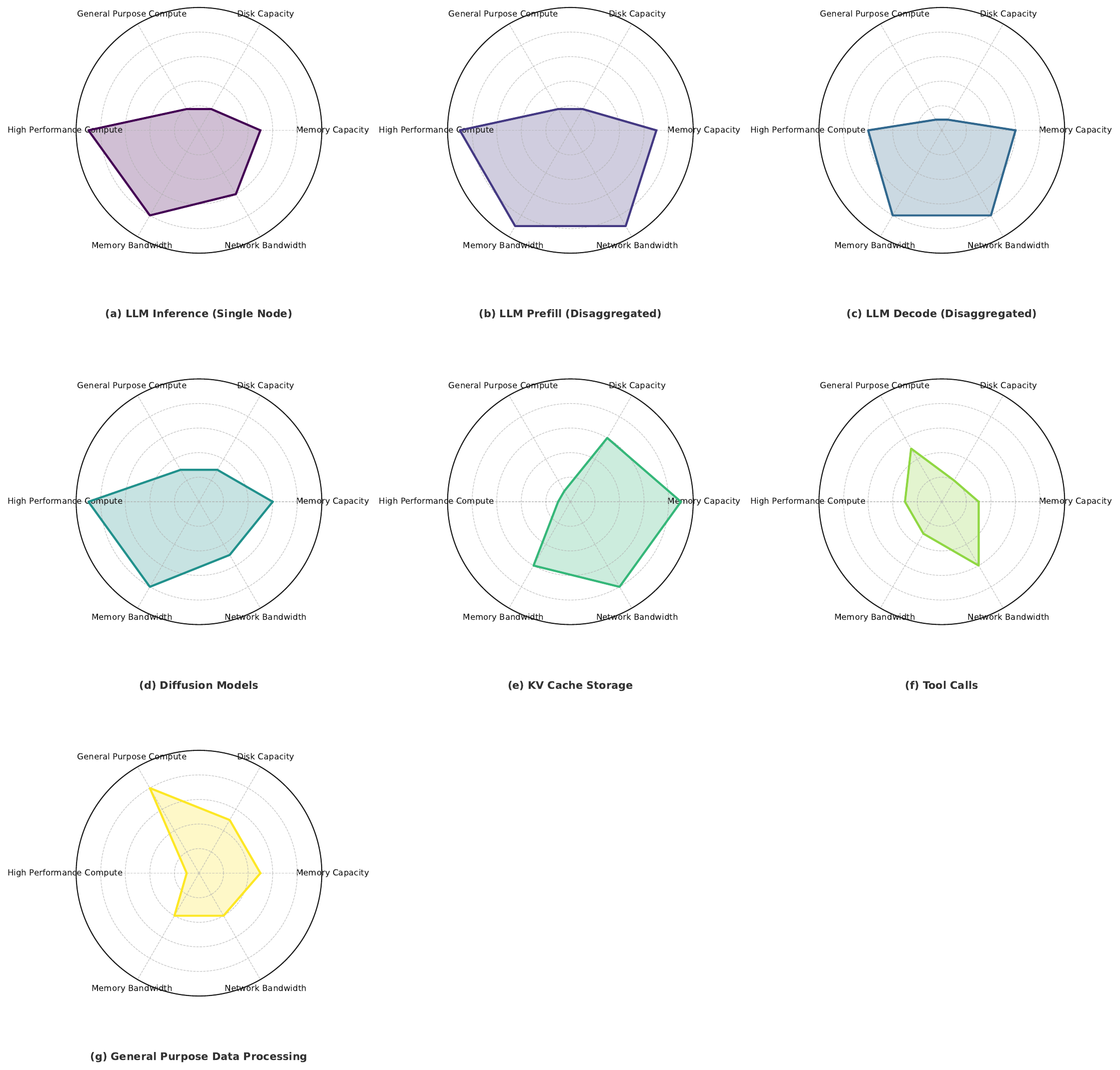}
    \caption{
        Radar plots comparing system resource demands across various AI workloads. Each subplot visualizes the relative importance (on a normalized scale from 0 to 10) of six key hardware dimensions: memory capacity, disk capacity, general purpose compute, high performance compute, memory bandwidth, and network bandwidth. These plots reflect qualitative estimates intended to illustrate workload characteristics, rather than results from direct performance measurements.
        \textbf{(a)}~LLM Inference (Single Node) is compute- and memory-intensive but operates within a single server, reducing network demands. 
        \textbf{(b)}~LLM Prefill (Disaggregated) requires high memory and network bandwidth due to distributed token processing. 
        \textbf{(c)}~LLM Decode (Disaggregated) has lower compute demand than prefill but still exhibits high memory and network usage. 
        \textbf{(d)}~Diffusion Models are broadly intensive across all dimensions, especially compute and memory bandwidth. 
        \textbf{(e)}~KV Cache Storage emphasizes memory and disk usage, with elevated network I/O for remote retrieval. 
        \textbf{(f)}~Tool Calls involve low compute but higher network bandwidth for accessing external tools or APIs. 
        \textbf{(g)}~General Purpose Data Processing is characterized by strong general-purpose compute and balanced use of disk, memory, and bandwidth.
    }
    \label{fig:workload-radar-charts}
\end{figure}

To illustrate how system requirements vary across different AI workloads, Figure \ref{fig:workload-radar-charts} presents radar plots of seven representative workload profiles. Each plot highlights the dominant hardware demands across critical system dimensions. We examine each workload in detail in Table \ref{tab:agent-workloads}.

\renewcommand{\arraystretch}{1.4}
\begin{table}[h]
\centering
\caption{Representative AI Agent Workloads and System Characteristics}
\label{tab:agent-workloads}
\begin{tabularx}{\textwidth}{l X}
\toprule
\textbf{Workload} & \textbf{Description} \\
\midrule
LLM Inference (Single Node) & This workload executes a transformer-based language model on a single machine to generate completions. It encompasses embedding, attention, feedforward layers, and output projection. The large matrix operations and continuous access to high-dimensional weight tensors result in high demand for compute and GPU memory capacity. Since execution is localized, network bandwidth requirements are negligible. Disk access is limited to model loading at initialization.
 \\[16pt]
LLM Prefill (Disaggregated) & The prefill phase processes the full input sequence to compute hidden states and populate the KV cache. It involves full attention across all tokens, requiring high compute throughput. Distributed execution amplifies demands on memory capacity, memory bandwidth, and network bandwidth.
 \\[16pt]
LLM Decode (Disaggregated) & During decode, the model generates one token per step using cached attention states. Although compute intensity is lower than prefill due to reduced matrix size, each step incurs frequent KV cache access. This results in sustained memory bandwidth usage. Depending on placement, cache accessibility also imposes significant demands on memory capacity or network bandwidth. \\[16pt]
Diffusion Models & Diffusion models iteratively transform noise into structured outputs over dozens to hundreds of inference steps. Each step involves a complete forward pass through a neural network, producing sustained and high compute utilization. The repeated loading of model parameters from high-bandwidth memory to on-chip SRAM across steps places sustained pressure on memory bandwidth, while intermediate storage is required for activations and state. \\[16pt]
KV Cache Storage & The KV cache holds layer-wise attention states for reuse during decoding. Long context windows increase memory footprint, while memory pressure can necessitate offloading to disk. In distributed settings, remote cache access introduces network overhead. Although compute is minimal, cache I/O latency is critical to maintaining low end-to-end response times under concurrent load. \\[16pt]
Tool Calls & Tool invocation involves calling external APIs or structured data sources. Since computation occurs externally, local compute and memory usage remain low. However, these steps introduce significant and variable network latency, as well as high outbound bandwidth requirements. General-purpose compute is needed for serializing requests, validating responses, and transforming results for downstream tasks. \\[16pt]
General Purpose Data Processing & This workload includes input/output formatting, control logic, and other auxiliary operations. Due to their general-purpose nature, These steps typically are typically executed on CPUs rather than specialized accelerators. Tasks such as document merging may require maintaining large in-memory buffers. Disk and network activity can arise when interfacing with external sources or storage systems. \\
\bottomrule
\end{tabularx}
\end{table}
\renewcommand{\arraystretch}{1}
\clearpage
\section{Design Framework for Heterogeneous Systems}

Today’s typical AI deployments rely heavily on racks of homogeneous GPUs interconnected via high-bandwidth, scale-up networks. These racks usually consist of identical nodes featuring high-performance GPUs such as NVIDIA’s GB200 NVL72\cite{nvidia2024blackwell}, each offering substantial computational resources (e.g., PFLOPS), large memory bandwidth (upwards of several TB/s), and substantial networking bandwidth facilitated by high-speed interconnects like NVIDIA's NVLink\cite{nvidia2025nvlink} or InfiniBand\cite{nvidia2025infiniband}. NVIDIA’s future roadmap further emphasizes scaling up these homogeneous rack-scale GPU systems to deliver even higher throughput and compute density, particularly targeting the demands of large-scale AI workloads \cite{nvidia2025hvdc}.

However, the uniformity of such homogeneous systems implies a uniform, high cost per unit of resource—whether it be computational FLOPs, memory bandwidth, network bandwidth, or memory capacity—regardless of the specific task or operation. Consequently, deployments incur substantial costs even for tasks that may not require the maximum available specifications in every dimension. For example, an asynchronous AI agent that relies on long-lived external requests may not benefit from a 10\% speedup in LLM inference, especially if it requires doubling infrastructure costs.

As highlighted previously, agentic AI workloads comprise diverse, dynamically orchestrated components with varying resource sensitivities. For instance, the decoding phase of large language models (LLMs) benefits significantly from greater memory capacity and bandwidth but does not demand as many computational FLOPs as the prefill stage. Similarly, data-intensive tasks might prioritize high memory bandwidth, while compute-bound operations require maximum computational power. Thus, using a homogeneous system forces a scenario where every task, irrespective of its specific resource requirements, incurs the same high uniform cost per unit of resource utilized.

\begin{figure}[htbp]
  \centering
  \includegraphics[width=\textwidth]{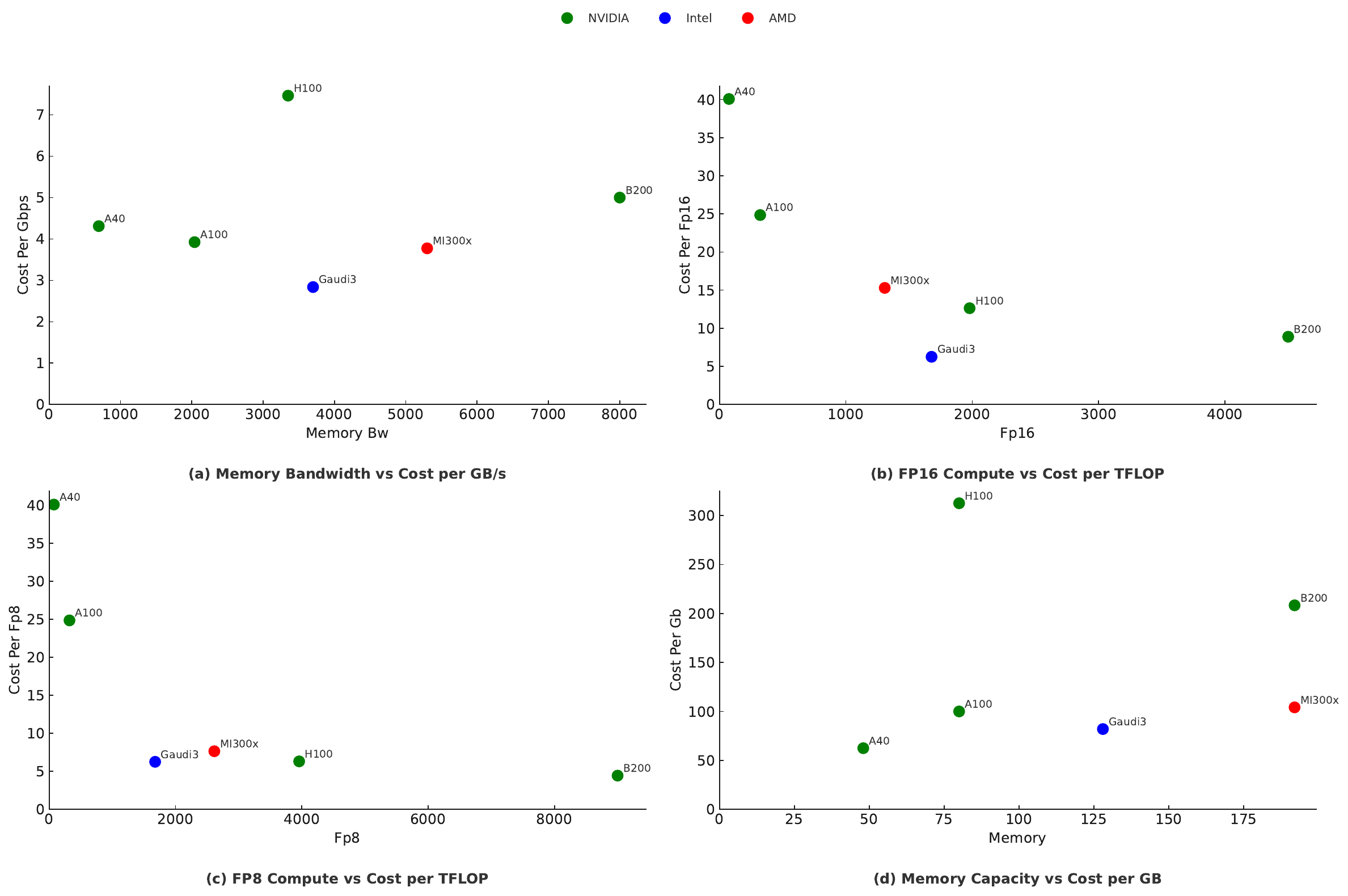}
  \caption{
    Marginal cost-efficiency analysis of contemporary AI accelerators, derived from publicly available hardware specifications~\cite{nvidia_a40, nvidia_a100, nvidia_h100, nvidia_b200, intel_gaudi3, amd_mi300x, nvidia_rtx6000}.
    Devices are color-coded by manufacturer: NVIDIA (blue), Intel (green), and AMD (red). 
    \textbf{(a)} Memory bandwidth versus cost per GB/s: Gaudi3 and MI300x exhibit the highest bandwidth efficiency. 
    \textbf{(b)} FP16 compute throughput versus cost per TFLOP: H100, Gaudi3, and MI300x provide strong cost-efficiency.
    \textbf{(c)} FP8 compute throughput versus cost per TFLOP: B200 offers leading efficiency at low precision.
    \textbf{(d)} Total memory capacity versus cost per GB: MI300x and A40 deliver the most cost-effective memory provisioning.
  }
  \label{fig:pareto-resource-tradeoffs}
\end{figure}

In contrast, heterogeneous systems offer a compelling alternative by integrating compute nodes (both accelerators and CPUs) with varied specifications across different resource dimensions such as TFLOPS, memory bandwidth and capacity, network bandwidth, disk capacity, and CPU resources. Figure.~\ref{fig:pareto-resource-tradeoffs} shows an analysis of the cost/unit for different resources for a sample of AI HW: memory BW, TFLOPS and memory capacity for different classes and generations of NVIDIA, Intel and AMD HW. As expected, different HW offer a diverse array of cost optimization points for different resources. By adopting such heterogeneous systems, individual components of an agentic AI graph could be optimally mapped onto specific hardware nodes tailored to their unique performance requirements. Consequently, instead of incurring a uniform cost across all operations, each component could leverage the most cost-effective hardware tailored to its needs, intuitively reducing the overall cost associated with executing agentic AI workloads.

The primary challenge in exploiting this insight lies in the lack of infrastructure for systemically representing, scheduling, and executing agentic AI workloads across heterogeneous hardware. In particular, doing so requires:
\begin{enumerate}
\item A fine-grained workload model that describes agent behavior as low-level execution steps (e.g., model inference, tool use, memory lookup).
\item A scheduling system that can reason about cost-performance trade-offs and assign each step to the most appropriate hardware, accounting for both compute and data transfer overhead.
\item A heterogeneous compiler and orchestration layer that abstracts away the complexities of generating and placing code on heterogeneous nodes, seamlessly integrating tasks using appropriate networking primitives, and ensuring compliance with SLAs. 
\end{enumerate}

When agent workloads are represented as task graphs, the assignment process becomes a constrained optimization problem over that structure. Under empirically grounded assumptions
based on hardware benchmarks, vendor specifications, and analytical models for resource usage, this reduces to a convex optimization problem—one that enables efficient and globally optimal planning across heterogeneous compute targets.

The next part formalizes this optimization framework. Subsequent parts in this section describe the systems-level mechanisms required to implement such a planner in practice — including profiling, cost modeling, and runtime execution across heterogeneous infrastructure.

\subsection{Formal Optimization Framework}
\label{sec:format-optimization}
With a structured representation of agent workloads as dataflow graphs, we can now formulate a principled optimization problem to assign tasks to heterogeneous hardware while minimizing cost and meeting performance constraints. These workloads include model execution, memory operations, retrieval and tool calls, as well as control logic — each with distinct compute and communication characteristics. In practice, this is often a multi-objective problem, where Pareto-optimal solutions must balance tradeoffs between cost, latency, energy, or other constraints.

We represent workloads as a directed graph \( G = (V, E) \), where each node \( i \in V \) is a task and each edge \( (i, k) \in E \) denotes a directed dependency. While many workloads are acyclic, our formulation supports general \emph{directed} graphs, including those with cycles (e.g., recurrent subgraphs, feedback loops), so long as execution constraints (like bounded unrolling or check-pointing) are satisfied in runtime planning.

Each task \( i \) must be assigned to a hardware class \( j \in H \) (e.g., GPU tier, CPU, accelerator). The challenge is to optimize this assignment across all tasks to minimize total system cost, subject to latency, capacity, and throughput constraints. The optimization framework is flexible and supports varying objective functions and constraint formulations, enabling adaptation to different system goals and deployment scenarios.

\subsubsection{Execution and Cost Model}

Each task consumes resources \( \theta_{ij}^{(r)} \) when executed on hardware \( j \), where \( r \in \{\text{compute}, \text{memory}, \text{bandwidth}\} \). Each device class \( j \) offers performance \( \text{perf}_j^{(r)} \) and has resource capacity \( \text{cap}_j^{(r)} \). Execution time is assumed to be bottlenecked by the slowest critical resource, and further affected by static overheads (e.g., network latency, kernel launch time) and communication costs:

\[
t_{ij} = \max_r \left( \frac{\theta_{ij}^{(r)}}{\text{perf}_j^{(r)}} \right) + l_i + d_{ij} + \delta_{ij}
\]

Where:
\begin{itemize}
  \item \( l_i \): static latency for task \( i \),
  \item \( d_{ij} \): pipeline parallelism or inter-device communication cost,
  \item \( \delta_{ij} \): synchronization overhead from task-parallelism (e.g., all-reduce).
  \item \( \max_r \left( \frac{\theta_{ij}^{(r)}}{\text{perf}_j^{(r)}} \right) \): latency for the slowest task in the graph.
\end{itemize}

In practice, these latency terms can be profiled from system traces, benchmarks, or prior executions, rather than analytically modeled.

The cost of executing task \( i \) on hardware \( j \) is modeled as:

\[
\text{Cost}_{ij} = \sum_r \theta_{ij}^{(r)} \cdot c_j^{(r)} + \gamma \cdot d_{ij}
\]

Where:
\begin{itemize}
  \item \( c_j^{(r)} \): cost per unit of resource \( r \) (e.g., per TFLOP or GB transferred),
  \item \( \gamma \): weight on inter-device communication penalties.
\end{itemize}

\subsubsection{Optimization Objective and Constraints}

\paragraph{Decision Variables.}

Let \( x_{ij} \in [0, 1] \) be the fraction of task \( i \) assigned to hardware class \( j \). In most systems, \( x_{ij} \in \{0, 1\} \), but fractional assignment can represent workload splitting or soft allocation.

\paragraph{Objective.}

Minimize the total execution cost across all tasks:
\[
\min \sum_{i \in V} \sum_{j \in H} x_{ij} \cdot \left( \sum_r \theta_{ij}^{(r)} \cdot c_j^{(r)} + \gamma \cdot d_{ij} \right) + \lambda \sum_{i \in V} s_i
\]
 
The slack variable \( s_i \) represents the amount by which task \( i \)'s latency can exceed its SLA target. By incorporating \( s_i \) in the objective function with a penalty weight \( \lambda \), the optimizer is incentivized to minimize SLA violations unless they yield significant cost savings.
Setting \( \lambda \to \infty \) enforces hard constraints.

\paragraph{Constraints.}

\begin{enumerate}
  \item \textbf{Assignment:}
  \[
  \sum_{j \in H} x_{ij} = 1 \quad \forall i \in V
  \]

  \item \textbf{Latency (with soft SLA):}
  \[
  t_i = \sum_{j} x_{ij} \cdot t_{ij}, \quad t_i - s_i \leq T_{\text{SLA}}, \quad s_i \geq 0
  \]

  \item \textbf{Throughput (optional):}
  \[
  \sum_{i \in V} \frac{1}{t_i} \geq R
  \]

  \item \textbf{Hardware capacity:}
  \[
  \sum_{i \in V} x_{ij} \cdot \theta_{ij}^{(r)} \leq \text{cap}_j^{(r)} \quad \forall j, r
  \]

  \item \textbf{Feasibility:}
  \[
  x_{ij} \in [0,1] \quad \forall i, j
  \]
\end{enumerate}

This convex formulation allows modeling heterogeneous system cost-performance tradeoffs, either through analytical resource scaling or profiled latency/cost estimates.

\subsubsection*{Worked Example: Optimizing Prefill/Decode under SLA}

As a concrete instantiation, we'll explore an example of optimizing a task node that is an LLM execution. We can further decompose the LLM execution task as a task graph \( G = (V, E) \) where:
\begin{itemize}
  \item \( V = \{\text{prefill}, \text{decode}\} \),
  \item \( E = \{ (\text{prefill} \rightarrow \text{decode}) \} \)
\end{itemize}
We consider two device types available for running each task: HP (high performance) and CO (cost optimized). This example models a single inference request, consisting of one prefill followed by one decode phase, with no context reuse or multi-turn interaction.

We assume:
\begin{itemize}
  \item Prefill processes 1000 input tokens,
  \item Decode generates 500 output tokens,
  \item Latency and cost per device-task pair are known from profiling. We assume these numbers are collected under optimal conditions for throughput and utilization, including maximized batching efficiency.
  \item There is no task splitting, i.e., each device exclusively executes its task. 
\end{itemize}

\begin{table}[ht]
\centering
\begin{tabular}{lccc}
\toprule
\textbf{Task} & \textbf{Device Type} & \textbf{Latency (ms)} & \textbf{Cost (per token)} \\
\midrule
Prefill      & HP     & 80              & \$0.00008 \\
Prefill      & CO     & 130             & \$0.00005 \\
Decode       & HP     & 25              & \$0.00006 \\
Decode       & CO     & 30              & \$0.00002 \\
\midrule
KV Transfer (HP → CO) & --- & 10 & \$0.000005 per prefill token \\
\bottomrule
\end{tabular}
\caption{Hypothetical latency and cost (profiled) for each task-device combination. HP represents a high-performance but expensive node, and CO represents a cost optimized option.}
\label{tab:prefill-decode-costs-hp-co}
\end{table}

Let \( T_{\text{SLA}} = 120 \) ms,  which reflects a typical threshold for interactive user experiences (e.g., web search or chatbot responses). We evaluate valid assignments:

\paragraph{Option A: Prefill and decode on HP}
\[
t = 80 + 25 = 105 \text{ ms}, \quad \text{SLA satisfied}
\]
\[
\text{Cost} = 1000 \cdot 0.00008 + 500 \cdot 0.00006 = \boxed{\$0.11}
\]

\paragraph{Option B: Prefill on HP, decode on CO}
\[
t = 80 + 30 + 10 = 120 \text{ ms}, \quad \text{SLA satisfied}
\]
\[
\text{Cost} = 1000 \cdot 0.00008 + 500 \cdot 0.00002 + 1000 \cdot 0.000005 = \boxed{\$0.095}
\]

\paragraph{Option C: All on CO}
\[
t = 130 + 30 = 160 \text{ ms}, \quad \text{SLA violated}
\]
\[
\text{Cost} = 1000 \cdot 0.00005 + 500 \cdot 0.00002 = \boxed{\$0.07}
\]

\paragraph{Optimization Result:}

Given \( t_i \leq T_{\text{SLA}} \), Option C is infeasible. Option B achieves lower cost than Option A while satisfying latency constraints. Thus, the optimal assignment is:
\[
x_{\text{prefill}, \text{HP}} = 1, \quad x_{\text{decode}, \text{CO}} = 1
\]

This decision reflects a fundamental tradeoff: while high-performance devices may offer lower latency, they incur a higher cost per token. Strategic disaggregation enables more efficient resource allocation, reducing overall cost while meeting strict latency requirements.

Further, this example demonstrates how the optimization leverages device heterogeneity to minimize cost without violating SLAs. In larger graphs, this formulation generalizes to multi-node pipelines, branching agent flows, and cyclic controller-executor patterns — using profiled characteristics or analytical estimates for runtime planning. This extends beyond model execution tasks and includes workloads composed of tool calls, external APIs, and data processing.

\section{System Design}

Building upon the cost optimization insights derived from our earlier framework in Section~\ref{sec:format-optimization}, we propose a comprehensive system architecture designed to facilitate flexible, efficient, and dynamic inference across heterogeneous hardware. The system takes as input a dataflow graph, dynamically schedules its execution according to optimal cost configurations computed via our convex optimization approach, and supports diverse AI agent workloads, including dynamic control flows, chained models, and external tool invocations. Moreover, it remains adaptable to evolving hardware landscapes comprising both high and low-end accelerators from vendors such as NVIDIA, Intel, and AMD.

\subsection{Orchestration and Serving System}

\begin{figure}[ht]
\centering
\includegraphics[width=0.9\textwidth]{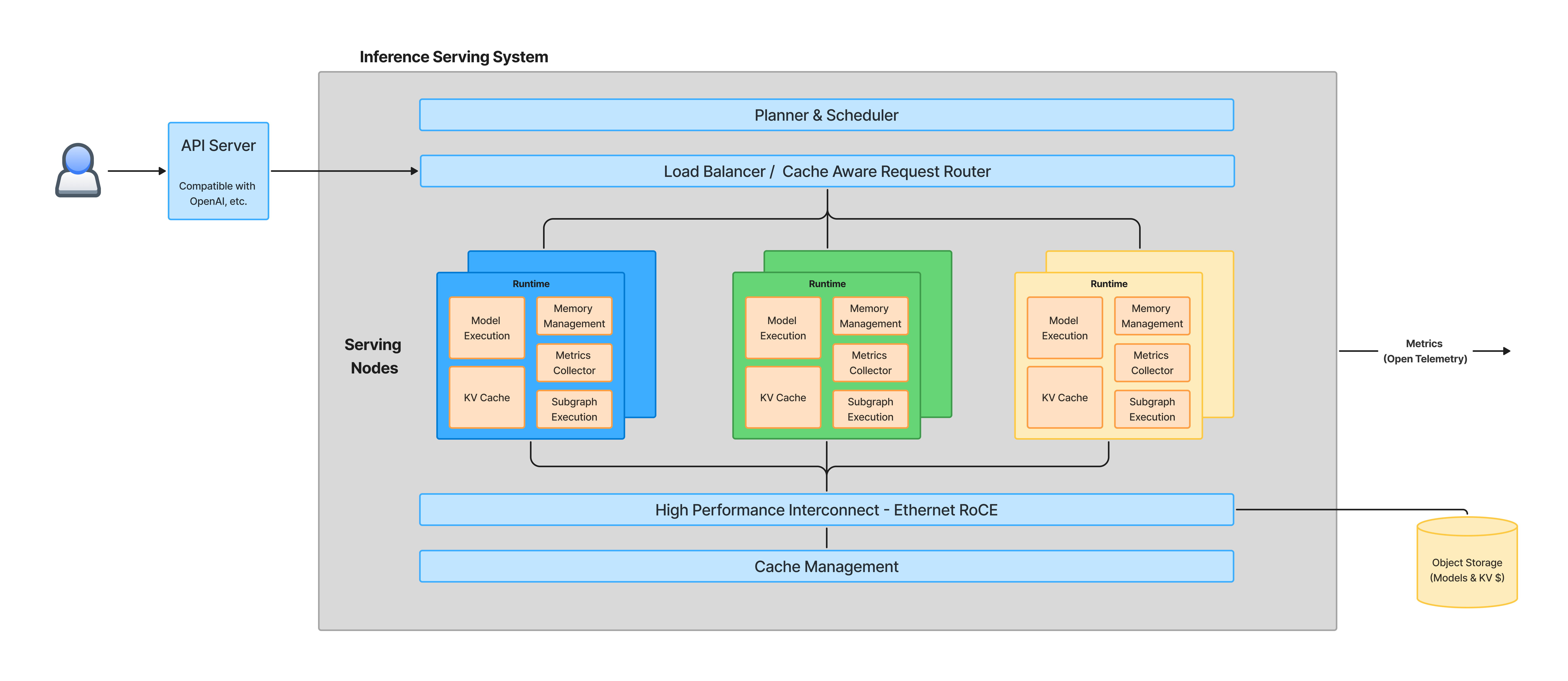}
\caption{High-level orchestration and serving system architecture}
\label{img:orchestration-system}
\end{figure}

Figure \ref{img:orchestration-system} illustrates the high-level design of our orchestration and serving system. The architecture prioritizes several key objectives:

\begin{enumerate}
\item \textbf{Scalability}: Automatically scales agentic workloads across heterogeneous hardware resources based on load and utilization.
\item \textbf{Flexibility}: Supports both synchronous workloads activated externally via APIs and asynchronous workloads operating autonomously.
\item \textbf{Composability}: Facilitates multi-turn interactions activated through repeated API calls or system state changes.
\end{enumerate}

The core functionality of our orchestration system involves dynamically planning and placing fine-grained computational components onto a distributed fleet of hardware. It continuously monitors node availability, workload characteristics, and resource utilization to inform placement decisions. This dynamic allocation helps prevent resource contention and optimizes both throughput and cost efficiency.

At a system level, the orchestration is designed to be deployed within distributed cluster environments, such as Kubernetes, employing a separation between a slow-path responsible for planning and resource allocation, and a fast-path for immediate execution.

The orchestration architecture includes the following primary components:

\begin{itemize}
\item \textbf{Planner \& Scheduler} (Slow Path): Continuously monitors hardware resources and workloads, dynamically allocating tasks based on the optimization strategies outlined in Section \ref{sec:format-optimization}. This component handles workload migration, resource allocation, and planning.

\item \textbf{Load Balancer / Request Router} (Fast Path): Routes requests based on cache locality and model availability, optimizing resource utilization and request aggregation for performance.

\item \textbf{Runtime}: Deployed to each accelerator node, the runtime encapsulates the core execution environment responsible for handling workloads from the scheduler. It implements mechanisms for model and subgraph execution, KV cache and memory management, and metrics collection. It is designed to run across heterogeneous environments by providing an abstraction to device specific capabilities. 

\item \textbf{RDMA Transport Layer}: Utilizes a high-performance Ethernet fabric with RDMA over Converged Ethernet (RoCE) to facilitate efficient transfer of models and caches. Abstraction layers and open standards ensure interoperability and optimal performance.

\item \textbf{Cache Manager}: Manages distributed key-value (KV) caches, graph databases, and agent memory storage, employing strategies for offloading less frequently accessed data to slower storage mediums such as secondary memory tiers, disks, or object storage.
\end{itemize}

For efficient execution across nodes, our system explicitly accounts for network topology and interconnect performance metrics, including latency, bandwidth, and potential contention. The orchestration system optimizes data communication patterns, particularly leveraging RoCE for workloads that require minimal latency between computational stages. Since KV cache placement significantly influences data movement, the cache management system actively coordinates distributed KV cache locations to minimize overheads associated with data transfers.

Running agentic graphs across heterogeneous environments presents unique challenges. These graphs are often not constructed with hardware diversity in mind, and naïvely executing them on heterogeneous systems can lead to inefficiencies or execution failures. To support intelligent scheduling and robust execution across varied accelerators, we must rethink how these workloads are represented and structured.

We propose the following components to address these requirements, as depicted in Figure \ref{fig:mlir-compiler-scheduler}:
\begin{itemize}
\item \textbf{Dataflow representation using MLIR dialect}: A MLIR dialect to represent the various components of an agentic AI graph to enable a systematic HW agnostic representation and enable planning.
\item \textbf{Dataflow orchestration system}: A scheduler that can analyze the workload, available hardware, and dynamically decide how to allocate granular fragments across hardware nodes.
\item \textbf{Dataflow compiler}: A flexible compiler capable of taking high level workloads, partitioning and translating nodes and operations into optimized kernels for different hardware backends.
\end{itemize}

\subsection{MLIR as a Foundation for Heterogeneous Agent Execution}

To support fine-grained analysis and optimization of agentic workloads across heterogeneous systems, we represent each workload as a program graph encoded in an intermediate representation. Specifically, we adopt the \textit{Multi-Level Intermediate Representation (MLIR)} framework \cite{lattner2021mlir}, which provides a rich, extensible infrastructure for expressing and transforming computations across abstraction levels.

\begin{figure}[ht]
\centering
\includegraphics[width=0.95\textwidth]{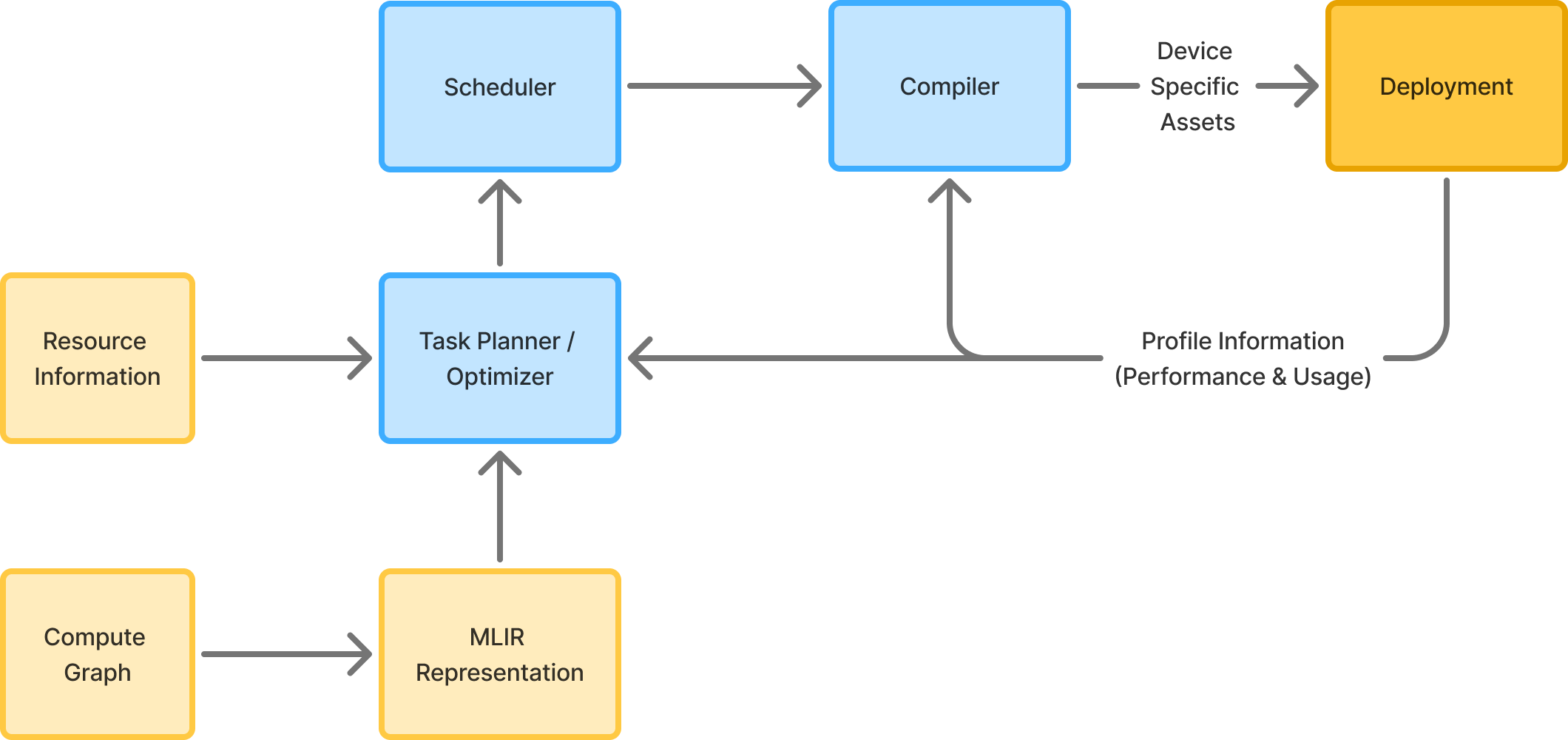}
\caption{System design stack from MLIR representation through task planning and compilation to deployment}
\label{fig:mlir-compiler-scheduler}
\end{figure}

Figure \ref{fig:mlir-compiler-scheduler} shows how MLIR acts as a bridge between the raw compute graph and deployment. Workloads are transformed through dialect-based intermediate representations, optimized using both static analysis and runtime resource feedback, and ultimately compiled and scheduled for execution across a heterogeneous set of backends.

Each task in the agent workload graph, ranging from LLM execution and memory access to tool calls and control logic, is assigned to an MLIR operation or subgraph. These operations are structured into dialects, which can be tailored to specific semantics (e.g., LLM inference, key-value caching, external tool APIs), enabling both domain-specific reasoning and cross-layer optimization.

\subsubsection*{MLIR for Agentic Workload Planning}

The use of MLIR brings several benefits to system design and execution planning:

\begin{enumerate}
  \item \textbf{Compositional Representation:} Tasks can be hierarchically composed, allowing for encapsulation of complex behavior (e.g., an agent that delegates to sub-agents or models).
  \item \textbf{Fusion and Decomposition:} Using MLIR transformations, adjacent or dependent operations can be fused to reduce communication overhead, or decomposed to enable distributed execution.
  \item \textbf{Static Analysis for Scheduling:} MLIR supports passes for bufferization, shape inference, cost estimation, and dependency analysis. These passes enable extraction of resource usage vectors \( \theta_{ij}^{(r)} \) and latency terms \( t_{ij} \), which feed directly into the convex optimization framework and scheduler.
  \item \textbf{Extensibility Across Modalities:} New dialects can capture novel task types (e.g., retrieval-augmented generation, multi-modal decoding), making the framework future-proof and adaptable to evolving agent designs.
  \item \textbf{Target-Aware Lowering:} Once optimized, MLIR graphs can be lowered into device-specific IRs such as LLVM IR~\cite{llvm}, TensorRT~\cite{tensorrt}, or XLA HLO~\cite{xla}, or alternate backends such as TVM~\cite{tvm}, IREE~\cite{iree}, or Glow~\cite{glow}, facilitating backend compilation for CPUs, GPUs, NPUs, and other accelerators.

\end{enumerate}

\subsubsection*{Agent frameworks to MLIR}

Modern agent frameworks such as LangChain allow users to express complex workflows using a high-level orchestration interface. These workflows typically compose memory, tool invocations, and language model calls in an imperative style. To optimize and schedule such workflows over heterogeneous hardware, we require a structured representation that exposes task boundaries, data dependencies, and operation semantics.

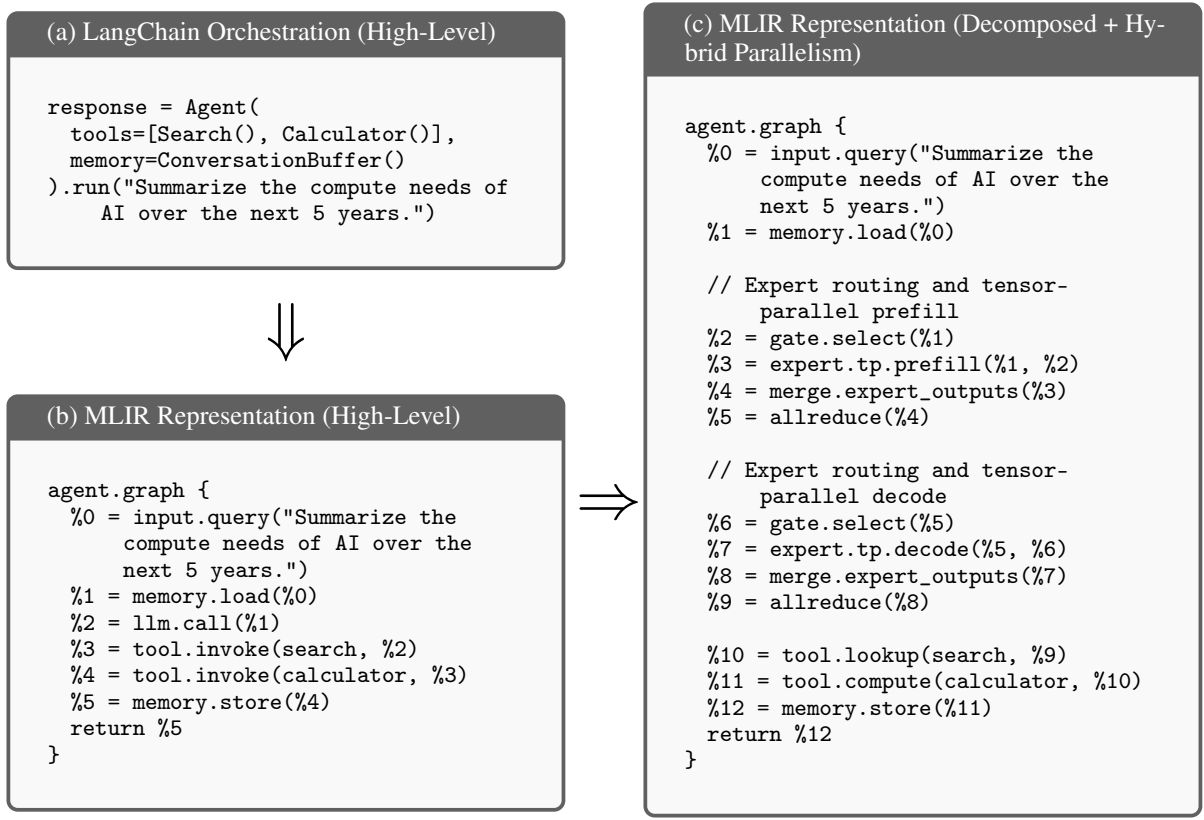
\begin{figure}[ht]
\centering
\begin{tikzpicture}[node distance=1em and 2em]

\node[inner sep=0pt, name=decomposed] (decomposed) {
\begin{minipage}[t]{0.45\textwidth}
\begin{tcolorbox}[colback=gray!5!white, colframe=gray!75!black, title={(c) MLIR Representation (Decomposed + Hybrid Parallelism)}]
\begin{lstlisting}
agent.graph {
  %0 = input.query("Summarize the compute needs of AI over the next 5 years.")
  %1 = memory.load(%0)

  // Expert routing and tensor-parallel prefill
  %2 = gate.select(%1)
  %3 = expert.tp.prefill(%1, %2)
  %4 = merge.expert_outputs(%3)
  %5 = allreduce(%4)

  // Expert routing and tensor-parallel decode
  %6 = gate.select(%5)
  %7 = expert.tp.decode(%5, %6)
  %8 = merge.expert_outputs(%7)
  %9 = allreduce(%8)

  %10 = tool.lookup(search, %9)
  %11 = tool.compute(calculator, %10)
  %12 = memory.store(%11)
  return %12
}
\end{lstlisting}
\end{tcolorbox}
\end{minipage}
};

\node[anchor=north west, name=langchain] at ($(decomposed.north west)+(-0.52\textwidth, 0)$) {
\begin{minipage}[t]{0.45\textwidth}
\begin{tcolorbox}[colback=gray!5!white, colframe=gray!75!black, title={(a) LangChain Orchestration (High-Level)}]
\begin{lstlisting}
response = Agent(
  tools=[Search(), Calculator()],
  memory=ConversationBuffer()
).run("Summarize the compute needs of AI over the next 5 years.")
\end{lstlisting}
\end{tcolorbox}
\end{minipage}
};

\node[below=4em of langchain, name=mlir] (mlir) {
\begin{minipage}[t]{0.45\textwidth}
\begin{tcolorbox}[colback=gray!5!white, colframe=gray!75!black, title={(b) MLIR Representation (High-Level)}]
\begin{lstlisting}
agent.graph {
  %0 = input.query("Summarize the compute needs of AI over the next 5 years.")
  %1 = memory.load(%0)
  %2 = llm.call(%1)
  %3 = tool.invoke(search, %2)
  %4 = tool.invoke(calculator, %3)
  %5 = memory.store(%4)
  return %5
}
\end{lstlisting}
\end{tcolorbox}
\end{minipage}
};

\node at ($(langchain.south)!0.5!(mlir.north)$) {\Huge$\Downarrow$};
\node at ($(mlir.east)!0.5!(decomposed.west)$) {\Huge$\Rightarrow$};

\node[fit=(langchain)(mlir)(decomposed), draw=none, inner sep=0pt] {};

\end{tikzpicture}

\caption{Transformation of a LangChain-style agent program into progressively lower-level MLIR representations. Panel (a) shows the original orchestration logic, while panels (b) and (c) illustrate how a compiler can lower this workflow into high-level and decomposed MLIR forms, enabling hybrid parallelism and heterogeneous hardware scheduling.}
\label{fig:mlir-transform-example}
\end{figure}

To optimize high-level agent workflows over heterogeneous infrastructure, the system must first lower these workflows into structured, semantically rich intermediate representations. Figure~\ref{fig:mlir-transform-example} illustrates this process. At the top, a LangChain-style orchestration defines an agent with memory and two tools—\texttt{Search()} and \texttt{Calculator()}—invoked in response to a user query. While simple to author, such imperative code lacks the structural annotations required for effective scheduling and device mapping.

The high-level MLIR representation (bottom left) introduces a typed dataflow graph over operations such as memory access, LLM invocation, and tool usage. On the bottom right, we show a decomposed variant that exposes internal parallelism within the model execution, enabling hardware-aware optimization. Specifically, we model a hybrid parallelism strategy that combines expert parallelism—where only a sparse subset of experts are activated per token—with tensor parallelism within each expert.

This is made explicit via a \texttt{gate.select} operation that routes input tokens to top-\(k\) experts. Each expert is then executed in parallel using \texttt{expert.tp.prefill} and \texttt{expert.tp.decode}, indicating a tensor-parallel subgraph per expert.

The MLIR representations below illustrate how such a workflow can be lowered into a structured, graph-based intermediate form. On the left, a high-level MLIR encoding captures typed operations such as memory load/store, LLM invocation, and tool usage. On the right, the same graph is decomposed into finer-grained operations: the LLM call is split into \texttt{prefill} and \texttt{decode}, and each tool invocation is separated into a \texttt{lookup} and a \texttt{compute} stage. This transformation reveals internal parallelism and resource requirements, enabling the compiler to reason about scheduling, placement, and pipelining across a heterogeneous system.

Each operation can be annotated with profiling metadata, resource usage estimates, or placement hints. A system pass then transforms this high-level IR into an annotated task graph ready for convex optimization.

\subsubsection*{Towards Heterogeneous-Aware Compilers}

The MLIR-based representation serves as a bridge between high-level workload semantics and low-level scheduling decisions. Just as traditional compilers optimize instruction placement and register allocation, this system-level compiler optimizes task placement and inter-device orchestration.

Crucially, such transformations are not limited to static programs. With extensions to support dynamic execution, asynchronous control flow, and runtime cost feedback, MLIR provides a foundation for compiling not just neural networks but distributed agentic systems that interact with tools, memory, and other agents in real time.

In this section, we have shown how we can start with a high-level agent definition that a developer might interact with, transform that into MLIR, and then use that MLIR to perform high-level optimizations. This optmized graph can then be scheduled using resource and run-time information across distributed infrastructure. Further, there is a compiler system that can take these MLIR operators and lower them to hardware-specific frameworks. Overall, this system allows for optimized deployment of entire agent workloads across a diverse set of hardware.

\section{Preliminary Results}

To assess the effectiveness of our system, we conducted an evaluation using a representative conversational voice agent composed of modular components: speech-to-text, text-to-speech, web search, and a central LLM node, as depicted in Figure~\ref{img:agent-dag}. These findings are preliminary, and comprehensive system validation is currently underway. Table~\ref{tab:gpu-specs} summarizes the accelerator hardware included in our current evaluation.

Our optimization framework places the non-LLM components of the voice agent on CPUs given the task characteristic (relatively computationally light) and the relative cost of a CPU, hence the dominant factor impacting overall TCO is the LLM component which is the most computationally demanding part. As a result, the following focuses on exploring optimizations on the LLM component. For the LLM, we evaluated four configurations of the LLaMA 3 model: 8B and 70B parameter sizes, each in FP16 and FP8 precisions (see Table~\ref{tab:model-specs}). Computational and memory demands were profiled based on model size, sequence lengths, and architectural details as an input to the optimization framework. Device-specific performance metrics, such as latency and throughput, incorporate empirical measurements when available and are augmented by theoretical roofline modeling~\cite{williams2009roofline} to represent realistic performance boundaries. All reported FLOP values assume dense computation, without accounting for sparsity.

To precisely isolate scheduling and hardware allocation benefits, we simulated a continuous workload scenario with unconstrained hardware availability. We evaluate which heterogeneous configuration leads to the maximum throughput (tokens/sec by maximizing batch size) under two different scenarios with SLAs that correspond to interactive and offline usage scenarios:

\begin{itemize}
\item \textbf{Latency SLA (Interactive workloads):} Time-to-First-Token (TTFT)  250 ms, Token-to-Token (TBT)  20 ms.
\item \textbf{Throughput SLA (Offline workloads):} Maximize $tokens/s/\$$.
\end{itemize}

\begin{table}[h]
\centering
\begin{tabular}{lccc}
\toprule
\textbf{Model} & \textbf{Parameters (B)} & \textbf{Precision} & \textbf{Source} \\
\midrule
LLaMA 3 - 8B - FP16 & 8 & FP16 & Meta AI~\cite{grattafiori2024llama} \\
LLaMA 3 - 8B - FP8 & 8 & FP8 & Meta AI~\cite{grattafiori2024llama} \\
LLaMA 3 - 70B - FP16 & 70 & FP16 & Meta AI~\cite{grattafiori2024llama} \\
LLaMA 3 - 70B - FP8 & 70 & FP8 & Meta AI~\cite{grattafiori2024llama} \\
\bottomrule
\end{tabular}
\caption{Model configurations used in evaluation.}
\label{tab:model-specs}
\end{table}

\begin{table}[!h]
\centering
\begin{tabular}{lcccccc}
\toprule
\textbf{Device} & \textbf{Manufacturer} & \textbf{Cost (\$)} & \textbf{Memory (GB)} & \textbf{Bandwidth (GB/s)} & \textbf{TFLOPs (FP16)} &\textbf{Operating Cost (\$/hr)} \\
\midrule
A40 & NVIDIA & \$3,000 & 48 & 696 & 75 & \$0.15\\
A100 & NVIDIA & \$8,000 & 80 & 2039 & 322 & \$0.25\\
Gaudi3 & Intel & \$12,500 & 128 & 3700 & 1678 & \$0.49 \\
MI300x & AMD & \$20,000 & 192 & 5300 & 1307 & \$0.52\\
H100 & NVIDIA & \$25,000 & 80 & 3350 & 1979 & \$0.60 \\
B200 & NVIDIA & \$40,000 & 192 & 8000 & 2250 & \$0.83 \\
\bottomrule
\end{tabular}
\caption{Specifications of accelerator hardware used in the optimizer. Costs averaged across a representative sample of hardware resellers available in public listings as of June 2025.}
\label{tab:gpu-specs}
\end{table}

\begin{figure}[!ht]
    \centering
    \includegraphics[width=\textwidth]{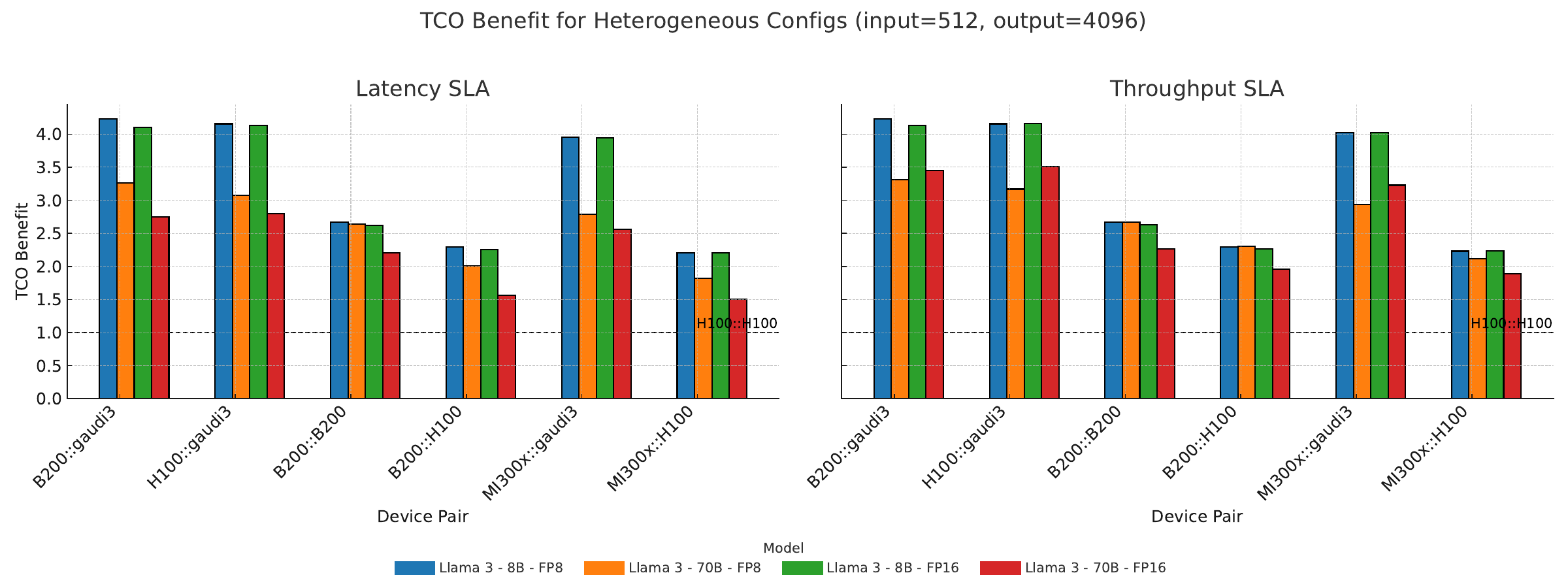}
    \caption{
        \textbf{TCO Benefit for Heterogeneous Configs (input=512, output=4096).}
        Comparison of cost efficiency across different Llama 3 models and device pairings. 
        Dashed line at 1.0 indicates baseline TCO for H100::H100. Bars show top configurations that meet SLA constraints:
        Latency SLA (TTFT $\leq$ 250ms, TBT $\leq$ 20ms) and Throughput SLA (Maximize $tokens/s/\$$).
        Results are based on a performance model fit to real measurements and explore heterogeneous configurations that leverage both tensor parallelism and pipeline parallelism with disaggregated inference.
    }
    \label{fig:tco-heterogeneous}
\end{figure}

\begin{figure}[!ht]
    \centering
    \includegraphics[width=\textwidth]{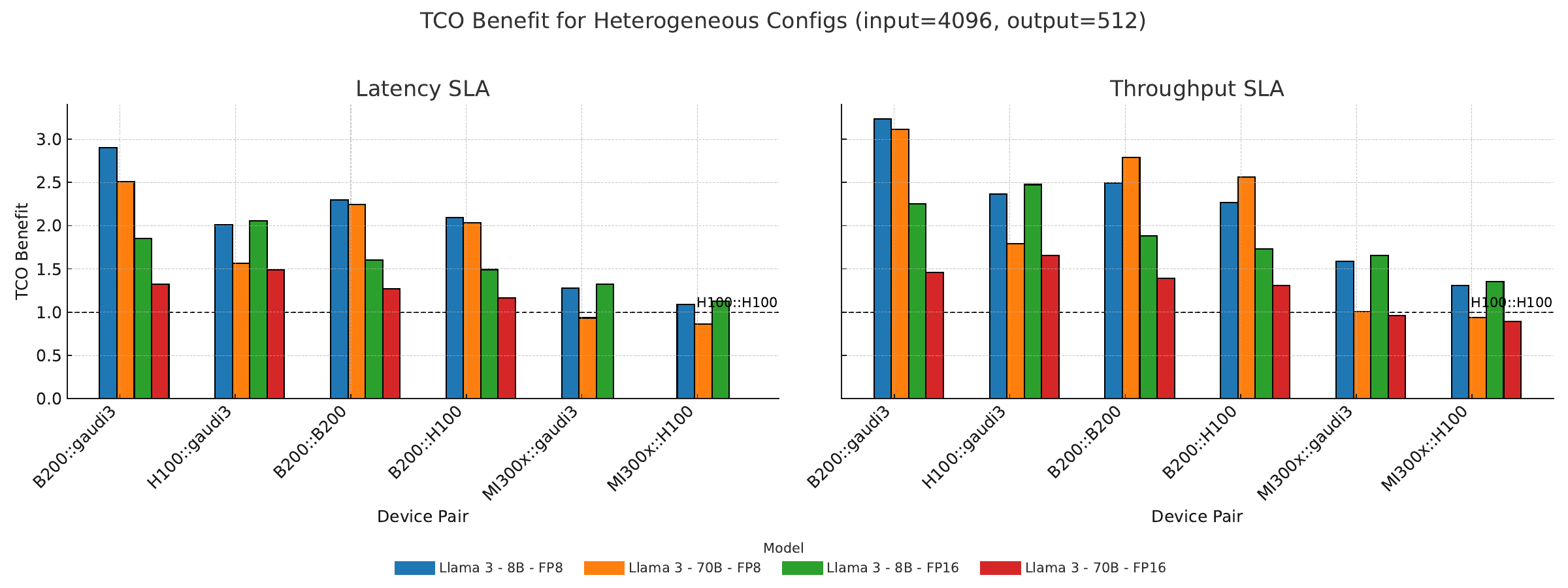}
    \caption{
        \textbf{TCO Benefit for Heterogeneous Configs (input=4096, output=512).}
        Comparison of cost efficiency across different Llama 3 models and device pairings. 
        Dashed line at 1.0 indicates baseline TCO for H100::H100. 
        Bars reflect top-performing configurations that satisfy SLA constraints:
        Latency SLA (TTFT $\leq$ 250ms, TBT $\leq$ 20ms) and Throughput SLA (Maximize $tokens/s/\$$.).
        These results are derived from a performance model calibrated to hardware measurements, incorporating both tensor parallelism and pipeline parallelism under disaggregated inference.
    }
    \label{fig:tco-heterogeneous-4096in-512out}
\end{figure}


\subsection{TCO of Heterogeneous Systems}

The evaluated accelerator hardware specifications are detailed in Table~\ref{tab:gpu-specs}, covering GPUs and ASIC accelerators from multiple vendors to demonstrate the broad applicability of our framework. The operating cost model assumes that hardware is financed over a fixed amortization period of 4 years with an interest rate of 8\%. For utility costs, we assume each node operates at its maximum rated TDP, with a cost of \$0.40/kWh. Other operational expenses, such as datacenter or colocation fees and nonrecurring engineering (NRE) costs, are excluded from the operating cost. Additionally, to describe the heterogeneous configurations, we leverage the operator "::" as a notation to denote disaggregated inference. The left and right operands correspond to the hardware configurations used during the prefill and decode stages, respectively.

Figures~\ref{fig:tco-heterogeneous}, and \ref{fig:tco-heterogeneous-4096in-512out} demonstrate the TCO improvements achievable through heterogeneous hardware configurations compared against the homogeneous baseline configuration (H100::H100). We focus on 6 possible combinations of hardware pairs, as they best illustrate the variations between performance and cost. We evaluated two scenarios for input-output sequence lengths, corresponding to reasoning tasks (long intermediate/output token sizes) and summarization tasks (short output sequence length). For each configuration, the system automatically explores options and selects the best combination of tensor and pipeline parallelism based on the available network bandwidth (both scale up and scale out) for that configuration and the latency SLA. Initial increases in tensor parallelism substantially reduced latency; however, further increases introduced significant device-to-device communication overhead, negating the computational efficiency gains. Additionally, our framework automatically incorporates optimizations such as paged attention \cite{kwon2023efficientmemorymanagementlarge}, further enhancing the efficiency of execution.

There are two interesting observations from the results.
\begin{itemize}
    \item \textbf{B200::Gaudi 3} has the best overall TCO benefit, especially for FP8 model configurations, for both interactive as well as batch workloads. The benefits are present (albeit smaller) even compared to a B200::B200 baseline which is the latest generation system. 
    \item \textbf{H100::Gaudi 3 configuration is often comparable or slightly better than a B200::B200 configuration}, implying that the Gaudi 3 can effectively complement the H100 and overall the heterogeneous configuration can deliver compelling performance, reducing the need to upgrade to Blackwell. The benefits are likely even higher if we incorporate the depreciation of the Hopper GPUs that have already been partially amortized, which is outside the scope of this paper. 
\end{itemize}

\subsection{Deployment requirements and considerations}
One of the central challenges in deploying workloads across distributed systems lies in managing the bandwidth and latency constraints imposed by the interconnect fabric linking accelerators. These interconnects are typically categorized as \emph{scale-up} or \emph{scale-out} fabrics. \emph{Scale-up fabrics} aim to deliver high-bandwidth, low-latency connections with shared memory semantics across multiple accelerators within a single system, as exemplified by NVLink-based designs such as NVL72~\cite{nvidia2024blackwell}. In contrast, \emph{scale-out fabrics} rely on commodity networking technologies such as Ethernet and InfiniBand~\cite{nvidia2025infiniband}, enabling the interconnection of large-scale clusters without shared memory, thereby requiring explicit software coordination for data movement.

In our system design, we assume that scale-up fabrics are confined to a single chassis, typically supporting up to 8 accelerators. Beyond this, we rely on high-speed \emph{RDMA over Converged Ethernet (RoCE)}~\cite{roceinitiative}, which is commonly deployed in modern large-scale AI datacenters~\cite{rdma2024}.

We utilize the underlying fabric for two primary purposes:
\begin{enumerate}[leftmargin=*]
    \item \textbf{Inter-node parallelism:} Distributing computation across multiple machines (for example tensor parallelism)
    \item \textbf{State transfer across pipeline stages:} Moving shared runtime state between nodes (for example key-value (KV) caches, during prefill/decode disaggregation.)
\end{enumerate}

Both inter-node parallelism and state transfer are incorporated into our total cost of ownership (TCO) model. The scalability of inter-node parallelism is constrained by the efficiency of data movement between accelerators, while state transfer primarily affects the end-to-end latency of the deployed agent.

Importantly, state transfer latency can often be partially amortized by overlapping communication with computation. For example, in prefill/decode disaggregation, key-value (KV) cache transfers contribute to the latency of the \textit{second token}, as the cache must be transmitted from the prefill stage to the decode stage. Fortunately, the bandwidth demands of this transfer are typically well-supported by modern AI datacenter networks~\cite{mitra2025beyond}. For completeness, we present the high-level bandwidth model that can be used to model the minimum bandwidth required to allow non-blocking pipelining of disaggregated inference:

\begin{equation}
BW_{\text{PeakEgress}} = \frac{\text{KV Cache Size}}{\text{TTFT} \cdot N_{PrefillGPU}}
\label{eq:peak_egress}
\end{equation}

\begin{equation}
BW_{\text{PeakIngress}} = \frac{\text{KV Cache Size}}{\text{TBT} \cdot N_{DeocodeGPU}}
\label{eq:peak_ingress}
\end{equation}

It is important to note that the above equations represent the \emph{peak} bandwidth required to transfer a single KV cache instance. In practice, inference systems often operate on batched inputs, which linearly scales the effective KV cache size and, correspondingly, the peak bandwidth requirement.

However, if the primary concern is overall task completion time—as is common in batch-oriented workloads—then it is more appropriate to consider \emph{amortized} bandwidth.

For practical workloads, we can estimate the peak bandwidth required based on the KV cache size and compute time. We compute the size of the key-value (KV) cache required for transformer-based models such as LLaMA using the following expression:

\begin{equation}
\text{KVCacheSize}_{\text{peak}} = 
2 \cdot N_{\text{layers}} \cdot d_{\text{model}} \cdot \left( \frac{N_{\text{kv}}}{N_{\text{heads}}} \right) \cdot \text{ISL} \cdot BS \cdot \text{BPE}
\label{eq:kv_cache_size_peak}
\end{equation}

\vspace{1em}
\noindent
\textbf{Legend:}
\begin{itemize}
    \item \( N_{\text{layers}} \): Number of transformer layers
    \item \( d_{\text{model}} \): Hidden dimension of the model
    \item \( N_{\text{kv}} \): Number of key/value heads
    \item \( N_{\text{heads}} \): Total number of attention heads
    \item \( \text{ISL} \): Input sequence length (tokens)
    \item \( BS \): Batch size
    \item \( \text{BPE} \): Bytes per element (e.g., 2 for \texttt{FP16})
\end{itemize}

Using the derived expressions, we observe that a 200--400~Gbps link is sufficient to meet the SLA requirements for transferring KV caches for input sequence lengths up to 32K tokens, depending on the specific LLaMA model variant employed. Such high-bandwidth interconnects are commonly available in modern high-performance AI datacenters.

While our TCO model incorporates a detailed treatment of networking latency and cost, we find that practical provisioning of interconnect bandwidth is generally sufficient to mitigate performance bottlenecks. Moreover, as noted by~\cite{mitra2025beyond}, increases in model and context size can actually reduce bandwidth requirements in practice. For instance, total time for first token (TTFT) tends to grow superlinearly with input sequence length (ISL), whereas the KV cache size grows only linearly. Similarly, while decode latency depends on the number of decoding GPUs, the corresponding ingress bandwidth requirement decreases inversely. Additionally, recent models with more efficient attention mechanisms—such as Multi-Linear Attention (MLA)—require smaller KV cache sizes~\cite{deepseekai2024deepseekv2strongeconomicalefficient}, further reducing pressure on interconnect bandwidth.

\subsection{Analysis}

To understand the above results, we explored how our optimization framework is making decisions on which parts of the voice agent workload are placed on which hardware. For example, the hardware allocations of different LLM inference stages (prefill and decode) are quite distinct given their different computational needs (prefill is computationally intensive whereas decode is more memory capacity intensive). Our framework inherently accommodates such optimizations by decomposing the LLM workload into granular components, enabling hardware resources to be matched precisely with operational demands.

Optimal hardware configurations varied significantly depending on input sequence length and decode tokens. For longer input sequences (Figure~\ref{fig:tco-heterogeneous-4096in-512out}), Intel Gaudi 3 accelerators emerged as the most cost-effective choice for prefill tasks due to their superior cost-performance ratio relative to NVIDIA B200. Conversely, when latency or FP8 performance is the primary concern, the higher computational power of the B200 justified its selection despite higher associated costs.

In decode-intensive scenarios (Figure~\ref{fig:tco-heterogeneous}), Gaudi3 accelerators were selected for decode tasks due to their lowest marginal cost, as indicated in Figure~\ref{fig:pareto-resource-tradeoffs}, assuming the workload can accommodate slightly longer token-to-token latency. Conversely, the B200 provides the best overall performance at an increased cost but remains relatively efficient compared to previous-generation systems such as the H100.

In conclusion, our optimization framework effectively leverages the diverse performance characteristics of heterogeneous hardware resources, dynamically allocating workloads based on specific SLA requirements. This adaptability enables optimal utilization of hardware capabilities, ensuring both cost efficiency and performance responsiveness tailored to individual requests.

\section{Related Work}
Recent advances in large-scale machine learning systems have led to the development of specialized infrastructure for model serving, disaggregated execution, compiler optimization, and multi-agent orchestration. In this section, we review representative work across each of these areas. While prior efforts offer important building blocks—ranging from low-level kernel optimization to high-level agent abstractions—they typically operate in isolation, without a unifying system that optimizes execution across heterogeneous compute. Our work builds on these foundations and introduces an optimization framework that integrates cost, performance, and hardware diversity into a cohesive planning model for AI agent workloads.

\subsection{Model Serving}

Recent advances in model serving have primarily targeted enhancing the efficiency and performance of Large Language Models (LLMs) through specialized software infrastructures. Prominent examples include the vLLM~\cite{kwon2023efficientmemorymanagementlarge} and TensorRT-LLM~\cite{NVIDIA2023tensorrtllm} frameworks, which have introduced significant software-level optimizations to enhance inference throughput, latency, and memory management.

vLLM introduces an innovative technique called paged attention, which substantially improves batched inference efficiency by effectively managing key-value (KV) caches. This design facilitates continuous batching, minimizes memory fragmentation, and is particularly suited for high-throughput, low-latency deployments. However, vLLM’s design is inherently model-centric and assumes a homogeneous hardware environment, thereby limiting its applicability to heterogeneous computing scenarios and comprehensive agentic workloads.

SGLang~\cite{sglang2024} represents a recent effort to provide a high-level programming interface for LLM serving, combining structured prompt orchestration with system-level performance optimizations. It incorporates a custom runtime and memory-aware scheduling to support latency-sensitive applications. However, like vLLM, SGLang primarily targets homogeneous infrastructure and single-model workloads, and does not address broader agentic or heterogeneous execution contexts.

TensorRT-LLM employs optimized CUDA kernels, quantization strategies, and operator fusion techniques to maximize GPU utilization. Specifically tailored to NVIDIA hardware, TensorRT-LLM achieves notable performance by closely aligning model structures with hardware-specific optimizations. However, this hardware-software tight coupling significantly restricts cross-vendor portability and flexibility.

In contrast to prior approaches focused primarily on maximizing throughput and minimizing latency within isolated runtime contexts, our research proposes a more generalized optimization framework that explicitly incorporates operational costs, hardware heterogeneity, and the comprehensive efficiency of entire AI agent workloads.

\subsection{Disaggregated Serving}

Recent studies have explored disaggregated inference architectures, where scheduling, execution, and memory management functionalities are decoupled and distributed across a heterogeneous set of computing resources.

Splitwise~\cite{patel2024splitwise} exemplifies this approach by explicitly decomposing inference workloads into prefill and decode stages, executed across distinct nodes. Splitwise also illustrates practical heterogeneous deployment by employing two different NVIDIA accelerators, selected based on distinct performance-cost trade-offs, demonstrating the potential efficiency benefits of adaptive resource allocation. 

NVIDIA's comprehensive inference stack, including NVIDIA Dynamo~\cite{NVIDIA2025dynamo}, provides an integrated solution designed explicitly for disaggregated inference workloads. Components such as NVIDIA Dynamo Planner, NVIDIA Dynamo Smart Router, NVIDIA Dynamo Distributed KV Cache Manager, and NVIDIA Inference Transfer Library (NIXL) address various stages from workload compilation and scheduling to execution. However, despite the stack’s completeness, it remains deeply embedded within NVIDIA’s proprietary hardware and software ecosystem, limiting its applicability to broader, vendor-neutral contexts.

The llm-d platform~\cite{llmd2025}, an extension of the vLLM framework, offers disaggregated inference by separating prefill and decode operations across individual nodes. Its scheduler determines optimal workload placement based on KV cache state, service-level agreements (SLAs), and system load. Nevertheless, llm-d inherits fundamental constraints from its vLLM foundation, notably restricting deployment to one model per node, which can limit efficient resource utilization.

Mitra et al.~\cite{mitra2025beyond} present an extensive empirical analysis of disaggregated inference, systematically evaluating numerous configurations across diverse workloads and hardware settings. Their findings highlight that disaggregated serving yields substantial benefits, particularly for workloads characterized by high prefill demands and larger model sizes. Moreover, they emphasize the necessity of dynamic rate matching and elastic resource scaling as critical strategies to achieve Pareto-optimal balances between throughput and interactivity.

Our optimization framework generalizes these approaches, integrating both disaggregated and monolithic serving strategies as specific instances within a unified optimization formulation. By explicitly considering cost, performance, and hardware heterogeneity, it facilitates effective optimization of AI agent workloads across diverse computational environments.

\subsection{MLIR-Based Efforts}

Several recent efforts leverage Multi-Level Intermediate Representation (MLIR) to optimize machine learning workloads across heterogeneous hardware. MLIR serves as a foundational tool enabling hardware-agnostic optimizations and transformations that facilitate efficient code generation for diverse computing architectures.

IREE~\cite{iree2022} and MHLO~\cite{mhlo2021} are prominent examples demonstrating MLIR's potential for portable, high-performance compilation. IREE supports comprehensive end-to-end compilation and execution, accommodating various backend targets including CPUs, GPUs, and accelerators. MHLO offers a standardized representation for tensor operations, streamlining the compilation and optimization pipeline across multiple hardware platforms. However, existing MLIR-based frameworks primarily target individual model execution and do not explicitly optimize across complex agentic workloads with disaggregated execution scenarios.

Triton~\cite{triton2022} represents a differentiated yet complementary approach to MLIR-based systems. Rather than exposing a general IR for graph-level transformations, Triton offers a Python-based programming model focused on writing highly efficient GPU kernels. Triton has been used effectively to optimize dense linear algebra and memory-bound kernels within LLM workloads, and integrates well with PyTorch through custom operations. However, Triton's scope is primarily focused on kernel-level optimization rather than end-to-end graph compilation, and lacks intrinsic mechanisms to target heterogeneous or disaggregated systems.

Our work similarly leverages MLIR but extends its use to optimize across entire agentic workloads, specifically addressing heterogeneous hardware and disaggregated execution contexts.

\subsection{Agent Frameworks}

A growing number of frameworks have emerged to structure, coordinate, and execute agentic workloads. LangGraph~\cite{langgraph2024multiagent} provides a graph-based programming model for composing agent behaviors as stateful transitions over tool and memory nodes, enabling fine-grained control over execution flow. CrewAI~\cite{crewai2024} and Autogen~\cite{autogen2023} introduce structured abstractions for collaborative multi-agent systems, with an emphasis on division of labor, role assignment, and tool integration. These systems facilitate modular composition of agents and streamline orchestration, though they often focus on the programming abstraction and rely on general-purpose runtimes.

In contrast, our work complements these abstractions by introducing a cost- and performance-aware execution planning layer. While existing frameworks provide high-level semantics for agent interaction, they do not address optimal task-to-hardware assignment or the underlying systems challenges associated with heterogeneous execution environments.

\section{Future Work}

This work opens numerous promising avenues for future research, particularly focused on expanding the capabilities and robustness of agentic workloads executed across heterogeneous computing environments. We identify two key directions below:

\subsection{Distributed Datacenter Scheduling and Optimization}

A crucial future research area involves developing sophisticated scheduling frameworks tailored for distributed datacenter environments. These environments typically span a range of capabilities, from smaller, geographically dispersed datacenters—such as enterprise-level facilities offering proximity and lower latency—to large-scale hyperscaler datacenters that provide significant computational power. Efficiently orchestrating agentic workloads across these heterogeneous datacenters requires advanced scheduling techniques that dynamically balance latency, resource availability, compute intensity, and cost considerations. Methods including hierarchical scheduling, intelligent workload migration, and dynamic load balancing~\cite{dean2013tail,chen2018distributed} could be explored to ensure optimal resource utilization and robust performance under varying conditions.

\subsection{Cross-Device Agent Planning (Cloud and Edge)}

Another important direction is extending our optimization framework to support seamless execution and planning of agentic tasks across hybrid cloud-edge deployments. Such cross-device planning must account for unique constraints like varying network latencies, data locality considerations, bandwidth variability, and heterogeneous compute capabilities between edge devices and cloud infrastructure. Recent protocols like Minion and MinionS~\cite{minions2025} demonstrate practical benefits of decomposing and parallelizing tasks between local and cloud language models, significantly reducing costs while preserving accuracy. Formalizing and generalizing these approaches into comprehensive optimization frameworks will enable rigorous guarantees and broader applicability. Developing efficient algorithms that can dynamically adapt task distribution in response to changing conditions (such as network disruptions, fluctuating workload characteristics, and evolving hardware availability) presents a compelling and non-trivial challenge. Research into these adaptive cross-device strategies will greatly enhance the applicability and resilience of distributed agent workloads in real-world deployments.

\subsection{Enabling AI Cloud Marketplace}

The orchestration system described in this paper can be built upon to enable the creation of a comprehensive AI cloud Platform-as-a-Service (PaaS) marketplace. Unlike proprietary AI marketplace offerings such as NVIDIA Lepton, our approach democratizes AI infrastructure access, allowing entities with spare AI compute capacity of whichever HW to participate actively in the marketplace. Such a marketplace would allow third-party developers and infrastructure providers to offer optimized, modular AI components and workflows, each precisely mapped onto cost-effective hardware configurations. By abstracting hardware complexities, the orchestration system facilitates seamless integration and dynamic scaling, enabling enterprises and individuals to easily discover, purchase, and deploy specialized AI services tailored to their specific requirements, thereby accelerating innovation and broadening AI adoption across diverse market segments.

\section{Conclusion}
This paper describes the motivation for as well as the design and implementation of a system for delivering efficient and scalable agentic AI over heterogeneous infrastructure. We believe that the AI revolution is still in its infancy, and systems innovation that can enable a distributed, scale-out AI infrastructure is critical to enable cost-effective scaling of AI. Our work is a first step in that direction.

\section*{Acknowledgments}
We thank James Bartlett for his contributions to early brainstorming and technical discussions. We are also grateful to Taras Sereda, Natalie Serrino, and Omid Azizi for their valuable feedback during the review process and for their thoughtful discussions. This work was supported in part by compute resources provided by Gimlet Labs, Inc.

\bibliographystyle{unsrt}  
\bibliography{0_main}

\end{document}